\documentclass[runningheads]{llncs}
\newcommand{\corr}{(\Letter)}

\usepackage{hyperref}

\usepackage{amsmath,amssymb,amsfonts}%
\usepackage{mathrsfs}
\usepackage{multirow}
\usepackage{tabularray}
\usepackage{xcolor}
\usepackage{subcaption}
\usepackage{float}
\usepackage{booktabs}
\usepackage{makecell}
\usepackage{graphicx}
\usepackage[misc]{ifsym}
\usepackage{mwe}

\usepackage{color}

\begin{document}

\title{Revisiting OmniAnomaly for Anomaly Detection: performance metrics and comparison with PCA-based models}

\titlerunning{Revisiting OmniAnomaly for Anomaly Detection}

\author{Bruna Alves\corr \and
Ana Martins \and
Armando J. Pinho \and
Sónia Gouveia\corr
}

\authorrunning{B. Alves et al.}

\institute{IEETA/DETI/LASI, University of Aveiro, 3810-193 Aveiro, Portugal \\
\email{\{bruna.alves, a.r.martins, ap, sonia.gouveia\}@ua.pt}}

\maketitle              

\begin{abstract}

Deep learning models have become the dominant approach for multivariate time series anomaly detection (MTSAD), often reporting substantial performance improvements over classical statistical methods. However, these gains are frequently evaluated under heterogeneous thresholding strategies and evaluation protocols, making fair comparisons difficult. This work revisits OmniAnomaly, a widely used stochastic recurrent model for MTSAD, and systematically compares it with a simple linear baseline based on Principal Component Analysis (PCA) on the Server Machine Dataset (SMD). Both methods are evaluated under identical thresholding and evaluation procedures, with experiments repeated across 100 runs for each of the 28 machines in the dataset. Performance is evaluated using Precision, Recall and F1-score at point-level, with and without point-adjustment, and under different aggregation strategies across machines and runs, with the corresponding standard deviations also reported. The results show large variability across machines and show that PCA can achieve performance comparable to OmniAnomaly, and even outperform it when point-adjustment is not applied. These findings question the added value of more complex architectures under current benchmarking practices and highlight the critical role of evaluation methodology in MTSAD research.

\end{abstract}

\section{Introduction}




Anomaly detection aims to identify observations that deviate significantly from the expected behavior of a system. In multivariate settings, anomalies are often characterized as instances that violate the dominant correlation structure of the data or that lie in low-density regions of the underlying distribution~\cite{Aggarwal2015,Alves2026}.
In the context of multivariate time series anomaly detection (MTSAD), the problem is particularly challenging due to high dimensionality, temporal dependencies, class imbalance and heterogeneous anomaly patterns.

Classical statistical approaches model normal behavior by estimating covariance structure and identifying deviations from a principal subspace, as in Principal Component Analysis (PCA)~\cite{Shyu2003}. More recently, deep learning models — such as recurrent neural networks and variational autoencoders — have been proposed to capture nonlinear relationships and temporal dynamics in the data. Among these, OmniAnomaly~\cite{Omni} is a widely adopted stochastic variational autoencoder (VAE)-based model that integrates recurrent architectures and latent state modeling to detect anomalies in multivariate time series.

Despite the growing popularity of deep generative and recurrent models for MTSAD, the empirical benefits of such architectural complexity are not always systematically validated against strong classical baselines. Recent research reports performance gains using increasingly complex models, yet differences in thresholding strategies, evaluation protocols and data preprocessing often make comparisons difficult. In particular, the widespread use of point-adjustment evaluation and optimal threshold selection may inflate reported performance, potentially obscuring the true contribution of the anomaly score model itself. Consequently, it remains an open question whether nonlinear and temporal models consistently outperform simpler linear approaches when evaluated under identical experimental settings. 

This study addresses this gap by conducting a systematic comparison between OmniAnomaly~\cite{Omni}, a stochastic VAE-based recurrent model, and a simple PCA-based baseline applied directly to standardized data~\cite{Shyu2003}. The comparison makes use the Server Machine Dataset (SMD), a widely adopted benchmark for MTSAD introduced together with OmniAnomaly~\cite{Omni}. Both methods are thus evaluated under identical thresholding strategies and with and without point-adjustment, enabling the isolation of three factors: the anomaly score model, the threshold selection procedure and the evaluation protocol. By analyzing performance at both global and machine levels, this work aims to assess whether temporal and nonlinear modeling provides a consistent advantage over linear correlation modeling in the SMD benchmark. Furthermore, by disentangling the three effects on performance (modeling, thresholding and evaluation), this study contributes to a more transparent assessment of anomaly detection methodologies for multivariate time series.

The remainder of this paper is organized as follows. Section~2 describes the SMD dataset and the anomaly detection models under comparison (OmniAnomaly vs PCA) as well as the study design and the evaluation protocol used in the study. Section~3 presents and discusses the results at both global and machine levels. Finally, Section~4 concludes the paper and discusses implications for future research in multivariate time series anomaly detection.

\section{Materials and Methods}

In this study, the SMD benchmark was used to compare OmniAnomaly and PCA. The models differ not only in complexity, with OmniAnomaly being a DL model with several layers of neural networks and PCA being a much simpler linear model, but also in the features they extract. 
According to Alves et al. (2026)~\cite{Alves2026}, OmniAnomaly is a nested model that models both temporal and spatial dependencies simultaneously by embedding a temporal model (gated recurrent units) in a spatial one (VAE). In contrast, PCA is a purely spatial model as it extracts linear correlations between variables but ignores the temporal structure of the data.

\subsection{Experimental Data}

The SMD dataset was collected from a large Internet company and made publicly available through the official OmniAnomaly GitHub repository~\cite{SMD_GitHub}. The dataset contains operational measurements from 28 distinct server machines, grouped into three subsets (1-, 2- and 3-).
Each machine has records of 38 key performance indicators that describe the system behavior, including metrics related to CPU load, memory usage, disk activity and network traffic~\cite{SMD_Features_GitHub}. The observations are equally spaced at 1-minute intervals and the set of MTS for each machine spans approximately five weeks of monitoring. 
For every machine, the data are divided into two equal-length parts: a training set containing normal behavior and a test 
set with labeled anomalies. 

Table \ref{tab:SMD_summary} presents statistics for the SMD dataset, including the total number of time points, anomaly points and anomaly segments. The servers exhibit substantial variability, with the percentage of anomaly points ranging from 0.21\% to 7.83\% and the number of anomaly segments per machine ranging from 1 to 30. This heterogeneity well justifies conducting a machine-level analysis rather than relying solely on aggregated performance metrics, even when global performance is summarized using the mean and standard deviation across machines, a practice that remains relatively uncommon in the MTSAD literature.


\begin{table}[p]
\centering
\caption{Summary for the SMD dataset: number of points (total, test, anomaly segments) and number/length of anomaly segments: min–max and mean (std).}
\label{tab:SMD_summary}
\begin{tabular}{c|c|c|c|c|c}
\toprule
\textbf{Machine} & \textbf{Time Points} & \textbf{Anomaly Points} & \multicolumn{3}{c}{\textbf{Anomaly Segments}} \\
 & \makecell[c]{All \\ [-1mm]\scriptsize(test)} & \makecell[c]{\# \\ [-1mm]\scriptsize(\% All)} & \# & Min--Max & \makecell[c]{Mean \\ [-1mm]\scriptsize(std)} \\
\midrule
1-1 & \makecell[c]{56958 \\ [-1mm]\scriptsize (28479)} & \makecell[c]{2694 \\ [-1mm]\scriptsize (4.73)} & 8 & 2--721 & \makecell[c]{336.75 \\ [-1mm]\scriptsize (272.45)} \\
1-2 & \makecell[c]{47388 \\ [-1mm]\scriptsize (23694)} & \makecell[c]{542 \\ [-1mm]\scriptsize (1.14)} & 10 & 3--156 & \makecell[c]{54.20 \\ [-1mm]\scriptsize (42.12)} \\
1-3 & \makecell[c]{47405 \\ [-1mm]\scriptsize (23703)} & \makecell[c]{817 \\ [-1mm]\scriptsize (1.72)} & 12 & 3--225 & \makecell[c]{68.08 \\ [-1mm]\scriptsize (70.76)} \\
1-4 & \makecell[c]{47413 \\ [-1mm]\scriptsize (23707)} & \makecell[c]{720 \\ [-1mm]\scriptsize (1.52)} & 12 & 3--205 & \makecell[c]{60.00 \\ [-1mm]\scriptsize (55.55)} \\
1-5 & \makecell[c]{47411 \\ [-1mm]\scriptsize (23706)} & \makecell[c]{100 \\ [-1mm]\scriptsize (0.21)} & 7 & 4--31 & \makecell[c]{14.29 \\ [-1mm]\scriptsize (8.83)} \\
1-6 & \makecell[c]{47377 \\ [-1mm]\scriptsize (23689)} & \makecell[c]{3708 \\ [-1mm]\scriptsize (7.83)} & 30 & 3--3161 & \makecell[c]{123.60 \\ [-1mm]\scriptsize (568.42)} \\
1-7 & \makecell[c]{47394 \\ [-1mm]\scriptsize (23697)} & \makecell[c]{2398 \\ [-1mm]\scriptsize (5.06)} & 13 & 3--1215 & \makecell[c]{184.46 \\ [-1mm]\scriptsize (378.17)} \\
1-8 & \makecell[c]{47397 \\ [-1mm]\scriptsize (23699)} & \makecell[c]{763 \\ [-1mm]\scriptsize (1.61)} & 20 & 3--371 & \makecell[c]{38.15 \\ [-1mm]\scriptsize (82.73)} \\
\midrule
2-1 & \makecell[c]{47387 \\ [-1mm]\scriptsize (23694)} & \makecell[c]{1170 \\ [-1mm]\scriptsize (2.47)} & 13 & 8--452 & \makecell[c]{90.00 \\ [-1mm]\scriptsize (142.41)} \\
2-2 & \makecell[c]{47399 \\ [-1mm]\scriptsize (23700)} & \makecell[c]{2833 \\ [-1mm]\scriptsize (5.98)} & 11 & 3--872 & \makecell[c]{257.55 \\ [-1mm]\scriptsize (369.88)} \\
2-3 & \makecell[c]{47377 \\ [-1mm]\scriptsize (23689)} & \makecell[c]{269 \\ [-1mm]\scriptsize (0.57)} & 10 & 3--91 & \makecell[c]{26.90 \\ [-1mm]\scriptsize (31.83)} \\
2-4 & \makecell[c]{47378 \\ [-1mm]\scriptsize (23689)} & \makecell[c]{1694 \\ [-1mm]\scriptsize (3.58)} & 20 & 3--401 & \makecell[c]{84.70 \\ [-1mm]\scriptsize (139.56)} \\
2-5 & \makecell[c]{47377 \\ [-1mm]\scriptsize (23689)} & \makecell[c]{980 \\ [-1mm]\scriptsize (2.07)} & 21 & 3--371 & \makecell[c]{46.67 \\ [-1mm]\scriptsize (96.52)} \\
2-6 & \makecell[c]{57486 \\ [-1mm]\scriptsize (28743)} & \makecell[c]{424 \\ [-1mm]\scriptsize (0.74)} & 8 & 3--118 & \makecell[c]{53.00 \\ [-1mm]\scriptsize (42.74)} \\
2-7 & \makecell[c]{47392 \\ [-1mm]\scriptsize (23696)} & \makecell[c]{417 \\ [-1mm]\scriptsize (0.88)} & 20 & 2--305 & \makecell[c]{20.85 \\ [-1mm]\scriptsize (65.53)} \\
2-8 & \makecell[c]{47405 \\ [-1mm]\scriptsize (23703)} & \makecell[c]{161 \\ [-1mm]\scriptsize (0.34)} & 1 & 161--161 & \makecell[c]{161.00 \\ [-1mm]\scriptsize (0.00)} \\
2-9 & \makecell[c]{57444 \\ [-1mm]\scriptsize (28722)} & \makecell[c]{1755 \\ [-1mm]\scriptsize (3.06)} & 10 & 2--414 & \makecell[c]{175.50 \\ [-1mm]\scriptsize (137.07)} \\
\midrule
3-1 & \makecell[c]{57400 \\ [-1mm]\scriptsize (28700)} & \makecell[c]{308 \\ [-1mm]\scriptsize (0.54)} & 4 & 21--131 & \makecell[c]{77.00 \\ [-1mm]\scriptsize (51.17)} \\
3-2 & \makecell[c]{47405 \\ [-1mm]\scriptsize (23703)} & \makecell[c]{1109 \\ [-1mm]\scriptsize (2.34)} & 10 & 3--837 & \makecell[c]{110.90 \\ [-1mm]\scriptsize (245.47)} \\
3-3 & \makecell[c]{47406 \\ [-1mm]\scriptsize (23703)} & \makecell[c]{632 \\ [-1mm]\scriptsize (1.33)} & 26 & 3--481 & \makecell[c]{24.31 \\ [-1mm]\scriptsize (91.47)} \\
3-4 & \makecell[c]{47374 \\ [-1mm]\scriptsize (23687)} & \makecell[c]{977 \\ [-1mm]\scriptsize (2.06)} & 8 & 3--786 & \makecell[c]{122.12 \\ [-1mm]\scriptsize (252.17)} \\
3-5 & \makecell[c]{47381 \\ [-1mm]\scriptsize (23691)} & \makecell[c]{426 \\ [-1mm]\scriptsize (0.90)} & 11 & 3--151 & \makecell[c]{38.73 \\ [-1mm]\scriptsize (58.95)} \\
3-6 & \makecell[c]{57452 \\ [-1mm]\scriptsize (28726)} & \makecell[c]{1194 \\ [-1mm]\scriptsize (2.08)} & 11 & 3--230 & \makecell[c]{108.55 \\ [-1mm]\scriptsize (83.57)} \\
3-7 & \makecell[c]{57410 \\ [-1mm]\scriptsize (28705)} & \makecell[c]{434 \\ [-1mm]\scriptsize (0.76)} & 5 & 7--311 & \makecell[c]{86.80 \\ [-1mm]\scriptsize (113.68)} \\
3-8 & \makecell[c]{57407 \\ [-1mm]\scriptsize (28704)} & \makecell[c]{1371 \\ [-1mm]\scriptsize (2.39)} & 6 & 17--573 & \makecell[c]{228.50 \\ [-1mm]\scriptsize (176.94)} \\
3-9 & \makecell[c]{57426 \\ [-1mm]\scriptsize (28713)} & \makecell[c]{303 \\ [-1mm]\scriptsize (0.53)} & 4 & 31--126 & \makecell[c]{75.75 \\ [-1mm]\scriptsize (39.09)} \\
3-10 & \makecell[c]{47385 \\ [-1mm]\scriptsize (23693)} & \makecell[c]{1047 \\ [-1mm]\scriptsize (2.21)} & 13 & 3--428 & \makecell[c]{80.54 \\ [-1mm]\scriptsize (126.16)} \\
3-11 & \makecell[c]{57391 \\ [-1mm]\scriptsize (28696)} & \makecell[c]{198 \\ [-1mm]\scriptsize (0.35)} & 3 & 6--126 & \makecell[c]{66.00 \\ [-1mm]\scriptsize (48.99)} \\
\midrule
\textbf{All} & \makecell[c]{1416825 \\ [-1mm]\scriptsize (708420)} & \makecell[c]{29444 \\ [-1mm]\scriptsize (2.08)}  & 327 & 2--3161 & \makecell[c]{90.04 \\ [-1mm]\scriptsize (238.42)} \\
\bottomrule
\end{tabular}
\end{table}

In this study, the data were provided to the model implementations as a multivariate sequence of the form 
\begin{equation}
\textbf{x}_{\bullet,t:t+w}= \begin{bmatrix}
\mathbf{x}_{\bullet,t} &\mathbf{x}_{\bullet,t+1} & \cdots & \mathbf{x}_{\bullet,t+w}    
\end{bmatrix}
\in \mathbb{R}^{M \times (w+1)},
\label{eq:window}
\end{equation}
where each vector $\mathbf{x}_{\bullet,t} \in \mathbb{R}^{M\times1} $ corresponds to the $M=38$ metrics observed at a given time $t=1, 2, \cdots$. 





\subsection{Anomaly Detection via OmniAnomaly}

\begin{figure}
    \centering
    \includegraphics[width=0.9\linewidth]{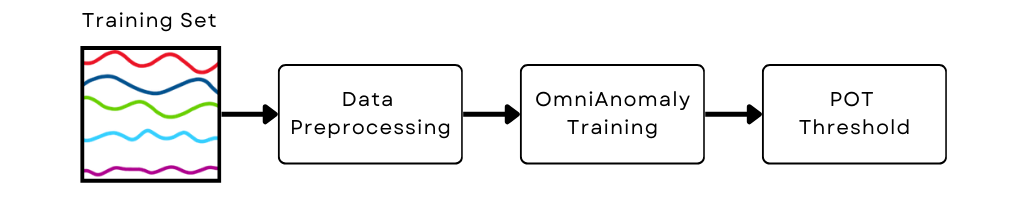}
    \caption{Pipeline for anomaly detection via OmniAnomaly. Adapted from~\cite{Omni}. }
    \label{fig:omni}
\end{figure}

Figure \ref{fig:omni} shows the OmniAnomaly anomaly detection pipeline~\cite{Omni}: data preprocessing, model training on normal observations, and threshold estimation using the Peaks-Over-Threshold (POT) method.
\vspace{-0.5cm}

The data preprocessing stage includes data scaling and windowing. 
The training and test sets are scaled separately by min-max scaling. Considering $\mathbf{x}_{\bullet,t} \in \mathbb{R}^{\text{M}}$ the original observation at time $t$, the standardized observation is defined as
\begin{equation}
\tilde{\mathbf{x}}_{\bullet,t} =
\left[
\frac{x_{1,t} - min_1}{max_1 - min_1},
\frac{x_{2,t} - min_2}{max_2 - min_2},
\ldots,
\frac{x_{M,t} - min_M}{max_M - min_M}
\right]^{\top},
\end{equation}
where $min_m$ and $max_m$, $m=1,\cdots, M$ are the minimum and maximum of metric $m$ computed over the respective set.

OmniAnomaly is a stochastic model that combines a variational autoencoder (VAE) with recurrent neural networks to capture temporal dependencies. Given a windowed input sequence (Eq.~\ref{eq:window}, $w=100$), the model consists of two main components: an inference network (encoder) and a generative network (decoder). Both networks incorporate Gated Recurrent Units (GRU) to model temporal dynamics. The encoder maps the input sequence to a latent representation $\mathbf{z}_t$, while the decoder reconstructs the observation conditioned on this latent state.

To further capture temporal structure in the latent space, OmniAnomaly integrates a linear Gaussian state-space model that allows the latent variables to evolve over time. In addition, planar Normalizing Flows (NF) are used to increase the flexibility of the posterior distribution through a sequence of invertible transformations. The model is trained by maximizing the Evidence Lower Bound (ELBO), which balances reconstruction accuracy and latent regularization. After training on normal data, anomaly scores are computed from the reconstruction likelihood. The anomaly score at time $t$ is defined as
\begin{equation}
AS_t = \log p_{\theta}(\tilde{\mathbf{x}}_{\bullet,t} \mid \mathbf{z}_{t-w:t}),
\end{equation}
where high scores indicate normal behavior (good reconstruction), while low scores suggest anomalies. 
An observation $\mathbf{x}_{\bullet,t}$ is considered anomalous when $AS_t$ falls below a threshold $th_F$, estimated using the POT method from Extreme Value Theory.
Since anomalies correspond to low reconstruction probabilities, the lower tail of the anomaly score distribution is modeled. Let $th$ denote an initial low quantile of the training scores. The exceedances
$th - AS \quad \text{for} \quad AS < th$ are assumed to follow a Generalized Pareto Distribution (GPD), whose survival function is given by
\begin{equation}
\bar{F}(s) = 
\mathbb{P}(th - AS > s \mid AS < th)
\sim
\left(1 + \frac{\gamma s}{\beta}\right)^{-1/\gamma},
\end{equation}
where $\gamma$ and $\beta$ denote the shape and scale parameters, respectively. These parameters are estimated via Maximum Likelihood Estimation using the training scores. The threshold $th_F$ is then computed as
\begin{equation}
th_F \approx
th -
\frac{\hat{\beta}}{\hat{\gamma}}
\left[
\left(\frac{q N'}{N'_{th}}\right)^{-\hat{\gamma}} - 1
\right],
\end{equation}
where $q$ is the desired tail probability, $N'$ is the total number of training observations and $N'_{th}$ is the number of scores satisfying $AS < th$.


\subsection{Anomaly detection via PCA}

PCA was implemented following the blocks in Figure \ref{fig:omni} and the procedures described in~\cite{Shyu2003}. The PCA requires a multivariate vector (Eq.~\ref{eq:window}, $w=0$), and therefore temporal dependencies are not explicitly modeled by this approach. Data preprocessing consists of standardization using statistics computed from the training set. The standardized observation is
\begin{equation}
\tilde{\mathbf{x}}_{\bullet,t} = 
\left[
\frac{x_{1,t} - \mu_1}{\sigma_1},
\frac{x_{2,t} - \mu_2}{\sigma_2},
\ldots,
\frac{x_{M,t} - \mu_M}{\sigma_M}
\right]^{\top},
\end{equation}
where $\mu_m$ and $\sigma_m$ are the mean and standard deviation of metric $m$ computed from the training data. The same standardization was applied to the test set. After standardization, the covariance structure of the training data is estimated. The sample covariance matrix is computed as
\begin{equation}
\mathbf{\Sigma} = \frac{1}{N} \sum_{t=1}^{N} 
\tilde{\mathbf{x}}_{\bullet,t} 
\tilde{\mathbf{x}}_{\bullet,t}^{\top},
\end{equation}
where $N$ is the number of training observations. Its eigenvalue decomposition is
\begin{equation}
\mathbf{\Sigma} = \mathbf{P}\mathbf{\Lambda}\mathbf{P}^{\top},
\end{equation}
where $\mathbf{P}$ contains the eigenvectors and $\mathbf{\Lambda}=\text{diag}(\lambda_1,\ldots,\lambda_M)$ the eigenvalues. Then, the anomaly score is computed as the sum of the deviations between the observation and its principal component projections, normalized by the explained variance
\begin{equation}
AS_t = \sum_{i=1}^{k} \frac{\left\|\tilde{\mathbf{x}}_{\bullet,t} - \mathbf{p}_i\right\|_2}{\text{EV}_i}
\end{equation}
where $\mathbf{p}_i$ is the $i$-th eigenvector and $\text{EV}_i$ is the explained variance ratio of the $i$-th principal component. Large values of $AS_t$ indicate that the observation lies far from the principal subspace describing normal behavior~\cite{Aggarwal2015,Shyu2003,PyOD_GitHub}. Finally, the number of components $k$ is selected using the cumulative explained variance
\begin{equation}
\text{CEV}(k) = \sum_{i=1}^{k}\text{EV}_i =
\frac{\sum_{i=1}^{k}\lambda_i}{\sum_{i=1}^{M}\lambda_i},
\end{equation}
choosing the smallest $k$ such that $\text{CEV}(k) \ge \tau$.

In addition to PCA, the arithmetic average was also considered as baseline model with the anomaly score is defined as
\begin{equation}\label{eq: mean}
AS_t = \left|\frac{1}{M}\sum_{m=1}^{M} \tilde{x}_{m,t} \right|,
\end{equation}
where $\tilde{x}_{m,t}$ is the standardized value of metric $m$ at time $t$. This baseline assigns equal weight to all metrics and ignores correlations between them. It is therefore an uninformative approach, as it requires no training data and does not model the underlying structure of normal data.



For consistency in the comparison with OmniAnomaly, the POT method was also applied to the PCA anomaly scores. Since PCA scores increase with the deviation from the principal subspace, POT was applied to the upper tail of the score distribution. Exceedances were defined as $AS - th$ for $AS > th$, and a GPD was fitted to model extreme values. Observations with scores exceeding the resulting threshold were classified as anomalies.


\subsection{Implementation}

The official OmniAnomaly implementation was used with the original hyperparameters~\cite{OmniAnomaly_GitHub,Omni}. The window length was set to $w=100$ and all other architectural and optimization parameters were kept unchanged from the default settings. Also, the initial threshold followed the original configuration using quantiles \{0.50\%, 0.75\%, 0.01\%\} for SMD groups \{1, 2, 3\}, with risk level $q=10^{-4}$.

PCA was implemented using the PyOD library~\cite{PyOD_GitHub,Zhao2019PyOD}, following the PCA-based anomaly detection formulation of~\cite{Shyu2003}. The number of retained components was selected using the cumulative explained variance criterion with $\tau = 0.5$. Although previous work used $k=20$ components~\cite{Audiber_thesis}, experiments with $\tau$ between 0.5 and 0.9 produced similar results, so $\tau$ was fixed at 0.5 for this study.


Regarding thresholding, POT was implemented using the code provided in the official OmniAnomaly repository~\cite{OmniAnomaly_GitHub}, with a minor modification in the sign convention to enable upper-tail modeling of PCA anomaly scores. For comparison, a Grid Search (GS) strategy was used to determine the optimal threshold for both OmniAnomaly and PCA. The original implementation suggests a GS range of [-400, 400] with step size 1 for machine 1-1~\cite{OmniAnomaly_GitHub}. However, this range was insufficient for all machines. Therefore, GS was performed over [-10000, 1000] with step size 1. For PCA, the number of thresholds was fixed to match that of OmniAnomaly and linearly spaced between the minimum and maximum 
anomaly scores in the test set. 
















\subsection{Study Design and Evaluation protocol}

The objective of this study is to compare OmniAnomaly with a simple linear PCA baseline under controlled and reproducible conditions. Both methods were evaluated on the same training and test splits of the SMD dataset. For each machine, models were trained exclusively on the training portion, assumed to contain only normal observations, without using anomaly labels. Anomaly scores were computed for all observations in the test set. Both methods were evaluated using the POT thresholding strategy implemented in the OmniAnomaly repository~\cite{OmniAnomaly_GitHub}, allowing the contribution of the anomaly scoring model to be isolated from that of threshold selection. For comparison, a Grid Search (GS) strategy was used to estimate an upper bound of achievable performance by evaluating all possible thresholds on the test set and selecting the one maximizing the F1-score.

Following prior work, performance was evaluated using Precision, Recall and F1-score:
\begin{equation}
P = \frac{TP}{TP + FP}, \quad R = \frac{TP}{TP + FN}, \quad F1 = \frac{2PR}{P + R},
\end{equation}
where $TP$, $FP$, $FN$ are true positives, false positives, and false negatives~\cite{ElAmineSehili2024,Omni}. Discrepancies in reported metrics may arise from how results are aggregated across $N=28$ machines and $r$ runs. Three strategies are considered: global average (A), macro-averaging (M), and micro-averaging (m).

Let $F1_{i,j}$ denote the F1-score for machine $i$ and run $j$. The averages across runs and machines are
\begin{align}
\overline{F1}_{i,\bullet} = \frac{1}{r}\sum_{j=1}^{r} F1_{i,j} \quad\text{ and } \quad \overline{F1}_{\bullet,j} = \frac{1}{N}\sum_{i=1}^{N} F1_{i,j} .\nonumber
\end{align}

The overall average (A) F1-score was computed equivalently as the average of the
machine means or of the run means i.e.
\begin{equation*}
\overline{F1}^{\text{A}} = \frac{1}{N}\sum_{i}\overline{F1}_{i,\bullet} = \frac{1}{r}\sum_{j}\overline{F1}_{\bullet,j},
\end{equation*}
with variability being reported using
\begin{equation*}
\sigma_{\text{mach.}} = \sqrt{\frac{1}{N-1}\sum_{i}\!\left(\overline{F1}_{i,\bullet}-\overline{F1}^{\text{A}}\right)^2} \text{ and }
\sigma_{\text{runs}} = \sqrt{\frac{1}{r-1}\sum_{j}\!\left(\overline{F1}_{\bullet,j}-\overline{F1}^{\text{A}}\right)^2},
\end{equation*}
to separate dataset heterogeneity across machines ($\sigma_{\text{mach.}}$) from the stochastic variability across runs ($\sigma_{\text{runs}}$).
For macro-averaging (M), precision and recall are averaged across machines,
\begin{equation}
\overline{P}_{\bullet,j} = \frac{1}{N} \sum_{i=1}^{N} P_{i,j}, \qquad
\overline{R}_{\bullet,j} = \frac{1}{N} \sum_{i=1}^{N} R_{i,j},
\end{equation}
and the aggregated score $F1_{\bullet,j}^{\text{M}}$ is computed from these averages, differing from $\overline{F1}_{\bullet,j}$, which averages machine-level F1-scores directly. In micro-averaging (m), $TP$, $FP$, $FN$ are are first aggregated across machine:
\begin{equation*}
P_{\bullet,j}^{\text{m}} = \frac{\sum_i TP_{i,j}}{\sum_i TP_{i,j} + \sum_i FP_{i,j}}, \qquad
R_{\bullet,j}^{\text{m}} = \frac{\sum_i TP_{i,j}}{\sum_i TP_{i,j} + \sum_i FN_{i,j}},
\end{equation*}
from which $F1_{\bullet,j}^{\text{m}}$ is obtained,
weighting machines proportionally to the number of detected events. Finally, the mean and standard deviation ($\sigma$) across runs were computed for both macro- and micro-averaged scores. 
Worst- and best-case performance was identified via hill-climbing over machine--run combinations~\cite{russell2010artificial}.

All metrics were computed at the point level after applying point-adjustment (PA), where an anomaly segment is considered detected if any point within it is flagged as anomalous~\cite{Omni,Wu2021}. Since PA tends to inflate Recall and F1-scores, results are also reported without PA using point-wise evaluation.


To complement point-level metrics, segment-level evaluation was also performed to characterize detection behavior within anomaly intervals. The number of detected episodes ($\mathcal{E}_d$) counts all threshold crossings into an anomaly state, while anomaly episodes ($\mathcal{E}_a$) count crossings restricted to true anomaly segments. The number of detected segments ($\mathcal{S}_d$) corresponds to the number of anomaly segments with at least one detected point. Finally, the anomaly length ($\mathcal{L}_a$) measures the percentage of detected points within each anomaly segment, for which minimum, maximum, mean, and standard deviation were computed. These metrics were also used to compare the detection behavior of OmniAnomaly and PCA beyond point-level accuracy.

\section{Results}

Table~\ref{tab:omnianomaly_smd} reports the performance of OmniAnomaly on the SMD dataset under POT and GS thresholding strategies. The upper panel reports previously published results~\cite{Omni}, showing that OmniAnomaly outperforms other baselines such as LSTM-NDT~\cite{Hundman2018}, DAGMM~\cite{Zong2018} and LSTM-VAE~\cite{Park2018}. The lower panel presents the results obtained in this study over $r=$100 independent runs using the same evaluation protocol.
All mean aggregated F1-scores are lower than those reported in Su et al.~\cite{Omni}, with the discrepancy being more pronounced under POT thresholding. The aggregation strategy that is most aligned with the results reported by Su et al.~\cite{Omni} is the macro averaging, reaching a maximum (best-case scenario) F1-score of 0.898 and 0.962 using POT and GS thresholds, respectively. There is, however, a non-negligible difference between the aggregation strategies that can be partly explained by class imbalance. Because anomalies are much rarer than normal observations, micro-averaging increases the influence of false positives, leading to lower precision and micro F1-scores. In contrast, macro-averaging assigns equal weight to each machine, while simple averaging computes the mean of machine-level F1-scores across runs. Finally, the standard deviations across runs exceed 1\% under POT thresholding, indicating that the POT-based threshold selection introduces additional variability in the resulting performance. In contrast, the GS strategy exhibits much lower variability across runs, as the optimal threshold is selected directly from the test anomaly scores. Overall, these results highlight the importance of considering multiple aggregation strategies when evaluating methods on the SMD dataset, as average F1-scores alone can mask substantial variability in detection performance across machines or runs.

\begin{table}[h!]
\centering
\caption{Performance on the SMD dataset. The upper panel shows previous results~\cite{Omni}. For GS, only F1-scores are available from Fig.~4 in~\cite{Omni} and were approximated via pixel-based extraction. The lower panel reports results in this study for 100 independent runs, as mean (std), minimum and maximum.\\}
\begin{tabular}{lccc|ccc}
\multicolumn{7}{c}{OmniAnomaly results (reported in Su et al. 2019~\cite{Omni})} \\
\hline
 & \multicolumn{3}{c|}{POT threshold} & \multicolumn{3}{c}{GS threshold} \\
 \hline
\textbf{Method} & \textit{P} & \textit{R} & \textit{F1} & \textit{P} & \textit{R} & \textit{F1} \\
\hline
LSTM-NDT~\cite{Hundman2018} & 0.568 & 0.644 & 0.604 & -- & -- & $\approx$ 0.68 \\
DAGMM~\cite{Zong2018} & 0.595 & 0.878 & 0.709 & -- & -- & $\approx$ 0.74 \\
LSTM-VAE~\cite{Park2018} & 0.792 & 0.708 & 0.784 & -- & -- & $\approx$ 0.80 \\
OmniAnomaly~\cite{Omni} & 0.833 & 0.945 & 0.886 & -- & -- & 0.962 \\
\hline\\
\multicolumn{7}{c}{OmniAnomaly results (100 runs, this study)} \\
\hline
 & \multicolumn{3}{c|}{POT threshold} & \multicolumn{3}{c}{GS threshold} \\
 \hline
\textbf{Evaluation} & $P$ & $R$ & $F1$ & $P$ & $R$ & $F1$ \\
\hline
Average &  0.855 & 0.772  & 0.746 & 0.949 & 0.926 & 0.933\\ 
\scriptsize $\sigma_{\text{runs}}$ & \scriptsize 0.016 & \scriptsize 0.015 & \scriptsize 0.015 & \scriptsize 0.009 & \scriptsize 0.010 & \scriptsize 0.005 \\
\scriptsize Min &\scriptsize 0.694 & \scriptsize0.722 & \scriptsize0.631 & \scriptsize0.936 & \scriptsize0.890 & \scriptsize0.903  \\
\scriptsize Max & \scriptsize 0.911 & \scriptsize 0.883 & \scriptsize 0.865 & \scriptsize 0.967 & \scriptsize 0.956 & \scriptsize 0.959 \\
\hline
Macro &  0.855 & 0.772 & 0.811 & 0.949 & 0.926 & 0.938\\ 
\scriptsize $\sigma_{\text{runs}}^{\text{M}}$ & \scriptsize 0.016 & \scriptsize 0.015 & \scriptsize 0.011 & \scriptsize 0.009 & \scriptsize 0.010 & \scriptsize 0.005 \\
\scriptsize Min & \scriptsize 0.706 & \scriptsize 0.707 & \scriptsize 0.706 & \scriptsize 0.920 & \scriptsize 0.895 & \scriptsize 0.907 \\
\scriptsize Max & \scriptsize 0.911 & \scriptsize 0.883 & \scriptsize 0.897 & \scriptsize 0.967 & \scriptsize 0.957 & \scriptsize 0.962 \\
\hline
Micro & 0.577  & 0.845  & 0.686  & 0.957 & 0.955 & 0.956 \\  
\scriptsize $\sigma_{\text{runs}}^{\text{m}}$ & \scriptsize 0.011 & \scriptsize 0.016 &  \scriptsize 0.010 & \scriptsize 0.013 & \scriptsize 0.008 &  \scriptsize 0.005 \\
\scriptsize Min & \scriptsize 0.509 & \scriptsize0.789 & \scriptsize 0.618 & \scriptsize 0.896 & \scriptsize 0.953 & \scriptsize 0.923  \\
\scriptsize Max & \scriptsize 0.638 & \scriptsize 0.927 & \scriptsize 0.756 & \scriptsize 0.975 & \scriptsize 0.976 & \scriptsize 0.975 \\
\hline
\end{tabular}
\label{tab:omnianomaly_smd}
\end{table}

To better understand the variability in detection performance across machines, Table~\ref{tab:results_omni_pa} reports the machine-level results of OmniAnomaly on the SMD dataset under the point-adjustment (PA) evaluation protocol. Performance is shown for both POT and GS thresholds for each machine individually. For most machines, the POT threshold yields F1, Precision, and Recall values close to those obtained with GS. However, for some machines (e.g., 1-5, 3-1, 3-3 and 3-4), POT appears to overestimate the threshold. As a result, anomalies with lower scores may not exceed the threshold and remain undetected, leading to lower Recall and F1-scores. The variability across machines is large (0.283) with some machines achieving near-perfect F1-scores (e.g., 1-1, 1-6) while others exhibit very low detection performance (e.g., 3-3, 3-4). Most machines show stable performance across runs, as indicated by the relatively small standard deviations. However, a few machines (e.g., 3-6, 3-8 and 3-10) exhibit higher variability across runs ($\geq 0.15$) which suggests that detection performance for these machines is more sensitive to variations in the anomaly scores and threshold selection. These results further highlight the importance of machine-level evaluation, since aggregate metrics alone may mask substantial differences in detection performance across machines.

Concerning the segment-level detection, the number of detected episodes ($\mathcal{E}_d$) is often larger than the number of anomaly episodes ($\mathcal{E}_a$), meaning that for most machines, there are detected episodes that are false detections, as they fall outside the ground truth anomaly segments. Moreover, the number of $\mathcal{E}_a$ is also often larger than the number of detected segments ($\mathcal{S}_d$), showing that OmniAnomaly does not detect each anomaly segment as a single contiguous episode. Instead, it flags several anomaly episodes within each ground truth anomaly segment. As an example, for machine-1-1, there are around 68 anomaly episodes for 7 detected segments. Furthermore, OmniAnomaly cannot detect the full length of each segment and typically identifies only a few points per segment. As an example, although it detects all three anomaly segments present in machine-3-11, it detects at most 16.67\% of each segment's total length. Overall, OmniAnomaly detects about 20.78\% of each segment's length across all machines. The point-adjustment strategy masks these aspects by considering an entire segment as detected the moment a single point falls within a true segment, which leads to overestimated detection performance.

\begin{table}[p]
\caption{Performance of OmniAnomaly using the POT and GS thresholding strategies under the point-adjustment (PA) evaluation protocol. Results are reported as mean (standard deviation) over 100 runs. Additional segment-level metrics include the number of detected episodes ($\mathcal{E}_d$), anomaly episodes ($\mathcal{E}_a$), detected segments ($\mathcal{S}_d$) and anomaly length ($\mathcal{L}_a$).}
\label{tab:results_omni_pa}
\centering
\resizebox{\textwidth}{!}{%
\begin{tabular}{c|cccccccc|cccccccc}
\toprule
\textbf{Machine} & \multicolumn{8}{c|}{\textbf{POT}} & \multicolumn{8}{c}{\textbf{GS}} \\
 & F1 & P & R & $\mathcal{E}_d$ & $\mathcal{E}_a$ & $\mathcal{S}_d$ & \multicolumn{2}{c|}{$\mathcal{L}_a$ (\%)} & F1 & P & R & $\mathcal{E}_d$ & $\mathcal{E}_a$ & $\mathcal{S}_d$ & \multicolumn{2}{c}{$\mathcal{L}_a$ (\%)} \\
 & & & & & & & \makecell[c]{mean \\ \scriptsize (std)} & \makecell[c]{max \\ \scriptsize (min)} & & & & & & & \makecell[c]{mean \\ \scriptsize (std)} & \makecell[c]{max \\ \scriptsize (min)} \\
\midrule
1-1 & \makecell[c]{0.985 \\ \scriptsize (0.008)} & \makecell[c]{0.972 \\ \scriptsize (0.016)} & \makecell[c]{0.999 \\ \scriptsize (0.001)} & \makecell[c]{110.84 \\ \scriptsize (31.86)} & \makecell[c]{67.73 \\ \scriptsize (11.99)} & \makecell[c]{7.12 \\ \scriptsize (0.73)} & \makecell[c]{15.53 \\ \scriptsize (17.68)} & \makecell[c]{50.00 \\ \scriptsize (0.00)} & \makecell[c]{0.999 \\ \scriptsize (0.000)} & \makecell[c]{1.000 \\ \scriptsize (0.000)} & \makecell[c]{0.997 \\ \scriptsize (0.000)} & \makecell[c]{8.81 \\ \scriptsize (2.04)} & \makecell[c]{8.81 \\ \scriptsize (2.04)} & \makecell[c]{5.00 \\ \scriptsize (0.00)} & \makecell[c]{0.37 \\ \scriptsize (0.74)} & \makecell[c]{3.97 \\ \scriptsize (0.00)} \\
1-2 & \makecell[c]{0.870 \\ \scriptsize (0.011)} & \makecell[c]{0.873 \\ \scriptsize (0.021)} & \makecell[c]{0.867 \\ \scriptsize (0.000)} & \makecell[c]{86.31 \\ \scriptsize (9.08)} & \makecell[c]{27.10 \\ \scriptsize (1.89)} & \makecell[c]{9.00 \\ \scriptsize (0.00)} & \makecell[c]{15.08 \\ \scriptsize (15.60)} & \makecell[c]{94.44 \\ \scriptsize (0.00)} & \makecell[c]{0.907 \\ \scriptsize (0.006)} & \makecell[c]{0.951 \\ \scriptsize (0.014)} & \makecell[c]{0.867 \\ \scriptsize (0.002)} & \makecell[c]{41.21 \\ \scriptsize (8.87)} & \makecell[c]{19.40 \\ \scriptsize (2.70)} & \makecell[c]{8.90 \\ \scriptsize (0.30)} & \makecell[c]{8.14 \\ \scriptsize (9.88)} & \makecell[c]{59.32 \\ \scriptsize (0.00)} \\
1-3 & \makecell[c]{0.914 \\ \scriptsize (0.007)} & \makecell[c]{0.963 \\ \scriptsize (0.009)} & \makecell[c]{0.870 \\ \scriptsize (0.012)} & \makecell[c]{81.32 \\ \scriptsize (11.23)} & \makecell[c]{58.17 \\ \scriptsize (7.81)} & \makecell[c]{8.49 \\ \scriptsize (0.58)} & \makecell[c]{18.23 \\ \scriptsize (16.66)} & \makecell[c]{50.52 \\ \scriptsize (0.00)} & \makecell[c]{0.929 \\ \scriptsize (0.010)} & \makecell[c]{0.968 \\ \scriptsize (0.031)} & \makecell[c]{0.895 \\ \scriptsize (0.043)} & \makecell[c]{60.88 \\ \scriptsize (38.62)} & \makecell[c]{42.47 \\ \scriptsize (24.11)} & \makecell[c]{9.01 \\ \scriptsize (1.06)} & \makecell[c]{16.05 \\ \scriptsize (16.70)} & \makecell[c]{50.52 \\ \scriptsize (0.00)} \\
1-4 & \makecell[c]{0.907 \\ \scriptsize (0.036)} & \makecell[c]{0.886 \\ \scriptsize (0.067)} & \makecell[c]{0.933 \\ \scriptsize (0.014)} & \makecell[c]{94.46 \\ \scriptsize (6.46)} & \makecell[c]{57.51 \\ \scriptsize (2.49)} & \makecell[c]{9.69 \\ \scriptsize (0.46)} & \makecell[c]{25.16 \\ \scriptsize (26.39)} & \makecell[c]{100.00 \\ \scriptsize (0.00)} & \makecell[c]{0.950 \\ \scriptsize (0.002)} & \makecell[c]{0.988 \\ \scriptsize (0.010)} & \makecell[c]{0.915 \\ \scriptsize (0.007)} & \makecell[c]{28.26 \\ \scriptsize (15.80)} & \makecell[c]{20.53 \\ \scriptsize (9.64)} & \makecell[c]{9.07 \\ \scriptsize (0.26)} & \makecell[c]{17.18 \\ \scriptsize (26.47)} & \makecell[c]{89.13 \\ \scriptsize (0.00)} \\
1-5 & \makecell[c]{0.241 \\ \scriptsize (0.080)} & \makecell[c]{0.583 \\ \scriptsize (0.079)} & \makecell[c]{0.154 \\ \scriptsize (0.065)} & \makecell[c]{11.19 \\ \scriptsize (2.47)} & \makecell[c]{1.78 \\ \scriptsize (0.73)} & \makecell[c]{1.74 \\ \scriptsize (0.63)} & \makecell[c]{4.41 \\ \scriptsize (10.15)} & \makecell[c]{50.00 \\ \scriptsize (0.00)} & \makecell[c]{0.729 \\ \scriptsize (0.076)} & \makecell[c]{0.651 \\ \scriptsize (0.082)} & \makecell[c]{0.846 \\ \scriptsize (0.128)} & \makecell[c]{47.81 \\ \scriptsize (17.72)} & \makecell[c]{6.96 \\ \scriptsize (1.48)} & \makecell[c]{6.01 \\ \scriptsize (0.80)} & \makecell[c]{22.47 \\ \scriptsize (25.00)} & \makecell[c]{80.00 \\ \scriptsize (0.00)} \\
1-6 & \makecell[c]{0.998 \\ \scriptsize (0.000)} & \makecell[c]{0.996 \\ \scriptsize (0.000)} & \makecell[c]{0.999 \\ \scriptsize (0.000)} & \makecell[c]{221.60 \\ \scriptsize (18.38)} & \makecell[c]{208.96 \\ \scriptsize (17.81)} & \makecell[c]{28.96 \\ \scriptsize (0.20)} & \makecell[c]{43.11 \\ \scriptsize (19.57)} & \makecell[c]{80.00 \\ \scriptsize (0.00)} & \makecell[c]{0.998 \\ \scriptsize (0.000)} & \makecell[c]{0.998 \\ \scriptsize (0.001)} & \makecell[c]{0.999 \\ \scriptsize (0.000)} & \makecell[c]{159.34 \\ \scriptsize (25.82)} & \makecell[c]{151.94 \\ \scriptsize (24.26)} & \makecell[c]{28.82 \\ \scriptsize (0.41)} & \makecell[c]{38.50 \\ \scriptsize (19.24)} & \makecell[c]{80.00 \\ \scriptsize (0.00)} \\
1-7 & \makecell[c]{0.968 \\ \scriptsize (0.012)} & \makecell[c]{0.942 \\ \scriptsize (0.024)} & \makecell[c]{0.996 \\ \scriptsize (0.006)} & \makecell[c]{167.93 \\ \scriptsize (60.85)} & \makecell[c]{34.13 \\ \scriptsize (5.62)} & \makecell[c]{12.51 \\ \scriptsize (0.69)} & \makecell[c]{40.68 \\ \scriptsize (30.95)} & \makecell[c]{99.92 \\ \scriptsize (0.00)} & \makecell[c]{0.987 \\ \scriptsize (0.004)} & \makecell[c]{0.984 \\ \scriptsize (0.006)} & \makecell[c]{0.991 \\ \scriptsize (0.008)} & \makecell[c]{62.52 \\ \scriptsize (15.80)} & \makecell[c]{26.99 \\ \scriptsize (3.99)} & \makecell[c]{11.93 \\ \scriptsize (0.82)} & \makecell[c]{33.33 \\ \scriptsize (31.53)} & \makecell[c]{99.92 \\ \scriptsize (0.00)} \\
1-8 & \makecell[c]{0.963 \\ \scriptsize (0.016)} & \makecell[c]{0.951 \\ \scriptsize (0.024)} & \makecell[c]{0.977 \\ \scriptsize (0.032)} & \makecell[c]{71.13 \\ \scriptsize (20.02)} & \makecell[c]{36.48 \\ \scriptsize (6.11)} & \makecell[c]{18.01 \\ \scriptsize (1.65)} & \makecell[c]{23.15 \\ \scriptsize (16.74)} & \makecell[c]{66.67 \\ \scriptsize (0.00)} & \makecell[c]{0.978 \\ \scriptsize (0.005)} & \makecell[c]{0.965 \\ \scriptsize (0.011)} & \makecell[c]{0.991 \\ \scriptsize (0.007)} & \makecell[c]{58.38 \\ \scriptsize (9.46)} & \makecell[c]{33.10 \\ \scriptsize (3.66)} & \makecell[c]{17.93 \\ \scriptsize (1.36)} & \makecell[c]{20.88 \\ \scriptsize (15.81)} & \makecell[c]{75.00 \\ \scriptsize (0.00)} \\
\midrule
2-1 & \makecell[c]{0.921 \\ \scriptsize (0.002)} & \makecell[c]{0.858 \\ \scriptsize (0.003)} & \makecell[c]{0.995 \\ \scriptsize (0.003)} & \makecell[c]{239.62 \\ \scriptsize (4.09)} & \makecell[c]{52.91 \\ \scriptsize (1.14)} & \makecell[c]{12.23 \\ \scriptsize (0.42)} & \makecell[c]{8.97 \\ \scriptsize (6.35)} & \makecell[c]{29.41 \\ \scriptsize (0.00)} & \makecell[c]{0.946 \\ \scriptsize (0.002)} & \makecell[c]{0.905 \\ \scriptsize (0.009)} & \makecell[c]{0.992 \\ \scriptsize (0.009)} & \makecell[c]{160.11 \\ \scriptsize (15.12)} & \makecell[c]{39.53 \\ \scriptsize (3.33)} & \makecell[c]{11.97 \\ \scriptsize (0.30)} & \makecell[c]{6.87 \\ \scriptsize (5.90)} & \makecell[c]{23.53 \\ \scriptsize (0.00)} \\
2-2 & \makecell[c]{0.263 \\ \scriptsize (0.009)} & \makecell[c]{0.151 \\ \scriptsize (0.006)} & \makecell[c]{1.000 \\ \scriptsize (0.000)} & \makecell[c]{1537.61 \\ \scriptsize (160.54)} & \makecell[c]{172.85 \\ \scriptsize (28.11)} & \makecell[c]{11.00 \\ \scriptsize (0.00)} & \makecell[c]{84.94 \\ \scriptsize (17.69)} & \makecell[c]{100.00 \\ \scriptsize (33.33)} & \makecell[c]{0.993 \\ \scriptsize (0.001)} & \makecell[c]{0.987 \\ \scriptsize (0.001)} & \makecell[c]{1.000 \\ \scriptsize (0.001)} & \makecell[c]{78.07 \\ \scriptsize (5.50)} & \makecell[c]{43.68 \\ \scriptsize (2.15)} & \makecell[c]{10.95 \\ \scriptsize (0.26)} & \makecell[c]{15.23 \\ \scriptsize (14.19)} & \makecell[c]{40.00 \\ \scriptsize (0.00)} \\
2-3 & \makecell[c]{0.964 \\ \scriptsize (0.013)} & \makecell[c]{0.930 \\ \scriptsize (0.024)} & \makecell[c]{1.000 \\ \scriptsize (0.000)} & \makecell[c]{39.01 \\ \scriptsize (8.03)} & \makecell[c]{20.78 \\ \scriptsize (2.86)} & \makecell[c]{10.00 \\ \scriptsize (0.00)} & \makecell[c]{35.94 \\ \scriptsize (22.77)} & \makecell[c]{82.93 \\ \scriptsize (7.41)} & \makecell[c]{1.000 \\ \scriptsize (0.001)} & \makecell[c]{0.999 \\ \scriptsize (0.001)} & \makecell[c]{1.000 \\ \scriptsize (0.000)} & \makecell[c]{14.41 \\ \scriptsize (1.10)} & \makecell[c]{14.23 \\ \scriptsize (0.91)} & \makecell[c]{10.00 \\ \scriptsize (0.00)} & \makecell[c]{27.60 \\ \scriptsize (21.62)} & \makecell[c]{80.00 \\ \scriptsize (1.23)} \\
2-4 & \makecell[c]{0.984 \\ \scriptsize (0.010)} & \makecell[c]{0.968 \\ \scriptsize (0.018)} & \makecell[c]{1.000 \\ \scriptsize (0.000)} & \makecell[c]{184.59 \\ \scriptsize (29.34)} & \makecell[c]{142.15 \\ \scriptsize (15.58)} & \makecell[c]{20.00 \\ \scriptsize (0.00)} & \makecell[c]{32.78 \\ \scriptsize (17.19)} & \makecell[c]{70.00 \\ \scriptsize (2.74)} & \makecell[c]{0.999 \\ \scriptsize (0.000)} & \makecell[c]{0.999 \\ \scriptsize (0.001)} & \makecell[c]{0.999 \\ \scriptsize (0.001)} & \makecell[c]{65.69 \\ \scriptsize (17.66)} & \makecell[c]{63.19 \\ \scriptsize (16.39)} & \makecell[c]{19.57 \\ \scriptsize (0.50)} & \makecell[c]{23.68 \\ \scriptsize (16.86)} & \makecell[c]{60.00 \\ \scriptsize (0.00)} \\
2-5 & \makecell[c]{0.769 \\ \scriptsize (0.023)} & \makecell[c]{0.626 \\ \scriptsize (0.031)} & \makecell[c]{1.000 \\ \scriptsize (0.000)} & \makecell[c]{332.31 \\ \scriptsize (25.65)} & \makecell[c]{74.84 \\ \scriptsize (7.22)} & \makecell[c]{21.00 \\ \scriptsize (0.00)} & \makecell[c]{42.42 \\ \scriptsize (19.19)} & \makecell[c]{86.24 \\ \scriptsize (4.85)} & \makecell[c]{0.937 \\ \scriptsize (0.011)} & \makecell[c]{0.948 \\ \scriptsize (0.046)} & \makecell[c]{0.931 \\ \scriptsize (0.055)} & \makecell[c]{74.11 \\ \scriptsize (42.51)} & \makecell[c]{36.08 \\ \scriptsize (9.49)} & \makecell[c]{19.57 \\ \scriptsize (0.50)} & \makecell[c]{25.97 \\ \scriptsize (15.58)} & \makecell[c]{63.38 \\ \scriptsize (0.00)} \\
2-6 & \makecell[c]{0.876 \\ \scriptsize (0.034)} & \makecell[c]{0.843 \\ \scriptsize (0.045)} & \makecell[c]{0.915 \\ \scriptsize (0.048)} & \makecell[c]{62.05 \\ \scriptsize (6.65)} & \makecell[c]{16.04 \\ \scriptsize (3.98)} & \makecell[c]{6.40 \\ \scriptsize (0.65)} & \makecell[c]{17.41 \\ \scriptsize (23.28)} & \makecell[c]{88.46 \\ \scriptsize (0.00)} & \makecell[c]{0.908 \\ \scriptsize (0.007)} & \makecell[c]{0.917 \\ \scriptsize (0.062)} & \makecell[c]{0.908 \\ \scriptsize (0.066)} & \makecell[c]{40.62 \\ \scriptsize (26.76)} & \makecell[c]{12.26 \\ \scriptsize (6.33)} & \makecell[c]{6.24 \\ \scriptsize (0.97)} & \makecell[c]{12.88 \\ \scriptsize (20.36)} & \makecell[c]{73.08 \\ \scriptsize (0.00)} \\
2-7 & \makecell[c]{0.740 \\ \scriptsize (0.036)} & \makecell[c]{0.592 \\ \scriptsize (0.046)} & \makecell[c]{0.990 \\ \scriptsize (0.000)} & \makecell[c]{177.06 \\ \scriptsize (17.98)} & \makecell[c]{28.65 \\ \scriptsize (3.82)} & \makecell[c]{19.00 \\ \scriptsize (0.00)} & \makecell[c]{33.16 \\ \scriptsize (11.69)} & \makecell[c]{66.67 \\ \scriptsize (0.00)} & \makecell[c]{0.978 \\ \scriptsize (0.004)} & \makecell[c]{0.977 \\ \scriptsize (0.009)} & \makecell[c]{0.978 \\ \scriptsize (0.007)} & \makecell[c]{25.50 \\ \scriptsize (3.30)} & \makecell[c]{18.56 \\ \scriptsize (0.92)} & \makecell[c]{17.95 \\ \scriptsize (0.63)} & \makecell[c]{25.17 \\ \scriptsize (12.54)} & \makecell[c]{50.00 \\ \scriptsize (0.00)} \\
2-8 & \makecell[c]{0.830 \\ \scriptsize (0.191)} & \makecell[c]{0.750 \\ \scriptsize (0.257)} & \makecell[c]{1.000 \\ \scriptsize (0.000)} & \makecell[c]{25.90 \\ \scriptsize (21.31)} & \makecell[c]{7.41 \\ \scriptsize (2.28)} & \makecell[c]{1.00 \\ \scriptsize (0.00)} & \makecell[c]{60.13 \\ \scriptsize (14.47)} & \makecell[c]{88.20 \\ \scriptsize (29.81)} & \makecell[c]{1.000 \\ \scriptsize (0.000)} & \makecell[c]{1.000 \\ \scriptsize (0.000)} & \makecell[c]{1.000 \\ \scriptsize (0.000)} & \makecell[c]{7.07 \\ \scriptsize (0.29)} & \makecell[c]{7.07 \\ \scriptsize (0.29)} & \makecell[c]{1.00 \\ \scriptsize (0.00)} & \makecell[c]{7.01 \\ \scriptsize (0.73)} & \makecell[c]{8.70 \\ \scriptsize (5.59)} \\
2-9 & \makecell[c]{0.968 \\ \scriptsize (0.010)} & \makecell[c]{0.944 \\ \scriptsize (0.005)} & \makecell[c]{0.994 \\ \scriptsize (0.020)} & \makecell[c]{273.94 \\ \scriptsize (22.55)} & \makecell[c]{200.92 \\ \scriptsize (16.26)} & \makecell[c]{9.91 \\ \scriptsize (0.29)} & \makecell[c]{32.30 \\ \scriptsize (18.50)} & \makecell[c]{66.67 \\ \scriptsize (0.00)} & \makecell[c]{0.978 \\ \scriptsize (0.005)} & \makecell[c]{0.958 \\ \scriptsize (0.009)} & \makecell[c]{0.999 \\ \scriptsize (0.007)} & \makecell[c]{192.75 \\ \scriptsize (52.82)} & \makecell[c]{138.52 \\ \scriptsize (40.40)} & \makecell[c]{9.99 \\ \scriptsize (0.10)} & \makecell[c]{25.95 \\ \scriptsize (21.57)} & \makecell[c]{66.67 \\ \scriptsize (0.00)} \\
\midrule
3-1 & \makecell[c]{0.595 \\ \scriptsize (0.031)} & \makecell[c]{0.972 \\ \scriptsize (0.007)} & \makecell[c]{0.429 \\ \scriptsize (0.041)} & \makecell[c]{7.83 \\ \scriptsize (0.90)} & \makecell[c]{5.01 \\ \scriptsize (0.10)} & \makecell[c]{1.01 \\ \scriptsize (0.10)} & \makecell[c]{0.96 \\ \scriptsize (1.65)} & \makecell[c]{3.82 \\ \scriptsize (0.00)} & \makecell[c]{0.895 \\ \scriptsize (0.004)} & \makecell[c]{0.969 \\ \scriptsize (0.010)} & \makecell[c]{0.831 \\ \scriptsize (0.000)} & \makecell[c]{13.15 \\ \scriptsize (2.71)} & \makecell[c]{6.00 \\ \scriptsize (0.00)} & \makecell[c]{2.00 \\ \scriptsize (0.00)} & \makecell[c]{1.15 \\ \scriptsize (1.57)} & \makecell[c]{3.82 \\ \scriptsize (0.00)} \\
3-2 & \makecell[c]{0.869 \\ \scriptsize (0.007)} & \makecell[c]{0.992 \\ \scriptsize (0.002)} & \makecell[c]{0.773 \\ \scriptsize (0.012)} & \makecell[c]{17.40 \\ \scriptsize (2.31)} & \makecell[c]{11.16 \\ \scriptsize (1.02)} & \makecell[c]{2.43 \\ \scriptsize (0.64)} & \makecell[c]{2.01 \\ \scriptsize (5.50)} & \makecell[c]{20.00 \\ \scriptsize (0.00)} & \makecell[c]{0.987 \\ \scriptsize (0.001)} & \makecell[c]{0.980 \\ \scriptsize (0.002)} & \makecell[c]{0.994 \\ \scriptsize (0.001)} & \makecell[c]{45.84 \\ \scriptsize (2.72)} & \makecell[c]{24.12 \\ \scriptsize (1.15)} & \makecell[c]{8.64 \\ \scriptsize (0.52)} & \makecell[c]{13.23 \\ \scriptsize (13.78)} & \makecell[c]{40.00 \\ \scriptsize (0.00)} \\
3-3 & \makecell[c]{0.065 \\ \scriptsize (0.083)} & \makecell[c]{0.983 \\ \scriptsize (0.025)} & \makecell[c]{0.037 \\ \scriptsize (0.076)} & \makecell[c]{1.66 \\ \scriptsize (0.83)} & \makecell[c]{1.32 \\ \scriptsize (0.58)} & \makecell[c]{1.32 \\ \scriptsize (0.58)} & \makecell[c]{0.52 \\ \scriptsize (3.03)} & \makecell[c]{33.33 \\ \scriptsize (0.00)} & \makecell[c]{0.951 \\ \scriptsize (0.007)} & \makecell[c]{0.975 \\ \scriptsize (0.010)} & \makecell[c]{0.929 \\ \scriptsize (0.013)} & \makecell[c]{35.63 \\ \scriptsize (5.61)} & \makecell[c]{23.78 \\ \scriptsize (1.35)} & \makecell[c]{22.63 \\ \scriptsize (0.69)} & \makecell[c]{23.24 \\ \scriptsize (13.61)} & \makecell[c]{60.00 \\ \scriptsize (0.00)} \\
3-4 & \makecell[c]{0.015 \\ \scriptsize (0.005)} & \makecell[c]{0.823 \\ \scriptsize (0.229)} & \makecell[c]{0.008 \\ \scriptsize (0.003)} & \makecell[c]{2.62 \\ \scriptsize (0.62)} & \makecell[c]{1.62 \\ \scriptsize (0.62)} & \makecell[c]{1.62 \\ \scriptsize (0.62)} & \makecell[c]{4.98 \\ \scriptsize (10.57)} & \makecell[c]{33.33 \\ \scriptsize (0.00)} & \makecell[c]{0.905 \\ \scriptsize (0.001)} & \makecell[c]{0.997 \\ \scriptsize (0.003)} & \makecell[c]{0.828 \\ \scriptsize (0.000)} & \makecell[c]{7.48 \\ \scriptsize (2.08)} & \makecell[c]{5.00 \\ \scriptsize (0.00)} & \makecell[c]{5.00 \\ \scriptsize (0.00)} & \makecell[c]{14.18 \\ \scriptsize (16.99)} & \makecell[c]{50.00 \\ \scriptsize (0.00)} \\
3-5 & \makecell[c]{0.755 \\ \scriptsize (0.009)} & \makecell[c]{0.996 \\ \scriptsize (0.003)} & \makecell[c]{0.608 \\ \scriptsize (0.012)} & \makecell[c]{13.82 \\ \scriptsize (3.65)} & \makecell[c]{12.77 \\ \scriptsize (2.98)} & \makecell[c]{5.26 \\ \scriptsize (1.61)} & \makecell[c]{10.65 \\ \scriptsize (14.50)} & \makecell[c]{33.33 \\ \scriptsize (0.00)} & \makecell[c]{0.772 \\ \scriptsize (0.004)} & \makecell[c]{0.985 \\ \scriptsize (0.012)} & \makecell[c]{0.635 \\ \scriptsize (0.006)} & \makecell[c]{23.16 \\ \scriptsize (4.79)} & \makecell[c]{18.93 \\ \scriptsize (1.94)} & \makecell[c]{8.55 \\ \scriptsize (0.70)} & \makecell[c]{19.83 \\ \scriptsize (14.08)} & \makecell[c]{33.33 \\ \scriptsize (0.00)} \\
3-6 & \makecell[c]{0.658 \\ \scriptsize (0.154)} & \makecell[c]{0.996 \\ \scriptsize (0.003)} & \makecell[c]{0.511 \\ \scriptsize (0.171)} & \makecell[c]{12.58 \\ \scriptsize (2.65)} & \makecell[c]{10.25 \\ \scriptsize (1.79)} & \makecell[c]{5.53 \\ \scriptsize (1.51)} & \makecell[c]{4.40 \\ \scriptsize (11.26)} & \makecell[c]{47.06 \\ \scriptsize (0.00)} & \makecell[c]{0.990 \\ \scriptsize (0.001)} & \makecell[c]{0.990 \\ \scriptsize (0.003)} & \makecell[c]{0.990 \\ \scriptsize (0.002)} & \makecell[c]{27.78 \\ \scriptsize (5.42)} & \makecell[c]{18.89 \\ \scriptsize (3.94)} & \makecell[c]{9.41 \\ \scriptsize (0.57)} & \makecell[c]{10.23 \\ \scriptsize (18.56)} & \makecell[c]{82.35 \\ \scriptsize (0.00)} \\
3-7 & \makecell[c]{0.914 \\ \scriptsize (0.041)} & \makecell[c]{1.000 \\ \scriptsize (0.000)} & \makecell[c]{0.845 \\ \scriptsize (0.068)} & \makecell[c]{3.81 \\ \scriptsize (1.14)} & \makecell[c]{3.81 \\ \scriptsize (1.14)} & \makecell[c]{2.68 \\ \scriptsize (0.66)} & \makecell[c]{4.12 \\ \scriptsize (5.25)} & \makecell[c]{14.29 \\ \scriptsize (0.00)} & \makecell[c]{1.000 \\ \scriptsize (0.000)} & \makecell[c]{1.000 \\ \scriptsize (0.001)} & \makecell[c]{1.000 \\ \scriptsize (0.000)} & \makecell[c]{6.93 \\ \scriptsize (0.79)} & \makecell[c]{6.74 \\ \scriptsize (0.69)} & \makecell[c]{5.00 \\ \scriptsize (0.00)} & \makecell[c]{6.76 \\ \scriptsize (4.22)} & \makecell[c]{14.29 \\ \scriptsize (1.59)} \\
3-8 & \makecell[c]{0.570 \\ \scriptsize (0.214)} & \makecell[c]{0.801 \\ \scriptsize (0.098)} & \makecell[c]{0.500 \\ \scriptsize (0.273)} & \makecell[c]{164.40 \\ \scriptsize (142.44)} & \makecell[c]{6.36 \\ \scriptsize (5.71)} & \makecell[c]{2.97 \\ \scriptsize (1.20)} & \makecell[c]{1.94 \\ \scriptsize (6.51)} & \makecell[c]{58.82 \\ \scriptsize (0.00)} & \makecell[c]{0.845 \\ \scriptsize (0.046)} & \makecell[c]{0.806 \\ \scriptsize (0.087)} & \makecell[c]{0.899 \\ \scriptsize (0.060)} & \makecell[c]{279.14 \\ \scriptsize (152.17)} & \makecell[c]{10.47 \\ \scriptsize (5.36)} & \makecell[c]{4.74 \\ \scriptsize (0.71)} & \makecell[c]{3.06 \\ \scriptsize (8.29)} & \makecell[c]{58.82 \\ \scriptsize (0.00)} \\
3-9 & \makecell[c]{0.746 \\ \scriptsize (0.083)} & \makecell[c]{0.764 \\ \scriptsize (0.139)} & \makecell[c]{0.760 \\ \scriptsize (0.112)} & \makecell[c]{91.07 \\ \scriptsize (72.70)} & \makecell[c]{11.16 \\ \scriptsize (3.45)} & \makecell[c]{2.23 \\ \scriptsize (0.62)} & \makecell[c]{3.72 \\ \scriptsize (5.37)} & \makecell[c]{39.68 \\ \scriptsize (0.00)} & \makecell[c]{0.846 \\ \scriptsize (0.028)} & \makecell[c]{0.945 \\ \scriptsize (0.055)} & \makecell[c]{0.768 \\ \scriptsize (0.041)} & \makecell[c]{22.03 \\ \scriptsize (16.48)} & \makecell[c]{8.57 \\ \scriptsize (1.32)} & \makecell[c]{2.18 \\ \scriptsize (0.39)} & \makecell[c]{2.49 \\ \scriptsize (3.52)} & \makecell[c]{19.35 \\ \scriptsize (0.00)} \\
3-10 & \makecell[c]{0.562 \\ \scriptsize (0.169)} & \makecell[c]{0.853 \\ \scriptsize (0.162)} & \makecell[c]{0.460 \\ \scriptsize (0.209)} & \makecell[c]{64.61 \\ \scriptsize (57.56)} & \makecell[c]{30.09 \\ \scriptsize (21.84)} & \makecell[c]{5.65 \\ \scriptsize (2.35)} & \makecell[c]{5.15 \\ \scriptsize (9.34)} & \makecell[c]{60.75 \\ \scriptsize (0.00)} & \makecell[c]{0.717 \\ \scriptsize (0.114)} & \makecell[c]{0.762 \\ \scriptsize (0.173)} & \makecell[c]{0.752 \\ \scriptsize (0.213)} & \makecell[c]{130.47 \\ \scriptsize (89.59)} & \makecell[c]{45.00 \\ \scriptsize (25.56)} & \makecell[c]{8.69 \\ \scriptsize (2.46)} & \makecell[c]{8.74 \\ \scriptsize (12.43)} & \makecell[c]{70.31 \\ \scriptsize (0.00)} \\
3-11 & \makecell[c]{0.965 \\ \scriptsize (0.020)} & \makecell[c]{0.934 \\ \scriptsize (0.036)} & \makecell[c]{1.000 \\ \scriptsize (0.000)} & \makecell[c]{25.58 \\ \scriptsize (9.59)} & \makecell[c]{11.18 \\ \scriptsize (0.61)} & \makecell[c]{3.00 \\ \scriptsize (0.00)} & \makecell[c]{10.03 \\ \scriptsize (4.79)} & \makecell[c]{16.67 \\ \scriptsize (5.56)} & \makecell[c]{0.988 \\ \scriptsize (0.002)} & \makecell[c]{0.976 \\ \scriptsize (0.004)} & \makecell[c]{1.000 \\ \scriptsize (0.000)} & \makecell[c]{15.22 \\ \scriptsize (1.10)} & \makecell[c]{10.39 \\ \scriptsize (0.71)} & \makecell[c]{3.00 \\ \scriptsize (0.00)} & \makecell[c]{9.24 \\ \scriptsize (5.31)} & \makecell[c]{16.67 \\ \scriptsize (3.03)} \\
\midrule
\textbf{Overall} & \makecell[c]{0.746 \\ \scriptsize (0.283)} & \makecell[c]{0.855 \\ \scriptsize (0.185)} & \makecell[c]{0.772 \\ \scriptsize (0.311)} & \makecell[c]{147.22 \\ \scriptsize (287.41)} & \makecell[c]{46.90 \\ \scriptsize (60.44)} & \makecell[c]{8.56 \\ \scriptsize (7.23)} & \makecell[c]{20.78 \\ \scriptsize (20.42)} & \makecell[c]{100.00 \\ \scriptsize (0.00)} & \makecell[c]{0.933 \\ \scriptsize (0.082)} & \makecell[c]{0.949 \\ \scriptsize (0.081)} & \makecell[c]{0.926 \\ \scriptsize (0.095)} & \makecell[c]{61.87 \\ \scriptsize (65.53)} & \makecell[c]{30.76 \\ \scriptsize (35.56)} & \makecell[c]{10.13 \\ \scriptsize (6.74)} & \makecell[c]{15.69 \\ \scriptsize (9.99)} & \makecell[c]{99.92 \\ \scriptsize (0.00)} \\
\bottomrule
\end{tabular}%
}
\end{table}


Turning now to the PCA baseline, the model shows competitive performance when compared to OmniAnomaly. Using the GS threshold, PCA achieves 
\begin{equation}
    \{\overline{F1}^{\text{A}}, \overline{P}, \overline{R}\} = \{0.930, 0.945, 0.935\},\nonumber
\end{equation}
with $\sigma_{\text{machines}} = 0.096$ and, under macro- and micro-averaging, the results are 
\begin{eqnarray}
    \{\overline{F1}^{\text{M}}, P^{\text{M}}, R^{\text{M}}\} &=& \{0.940, 0.945, 0.935\}\nonumber\\
    \{\overline{F1}^{\text{m}}, P^{\text{m}}, R^{\text{m}}\} &=& \{0.950, 0.930, 0.971\}.\nonumber
\end{eqnarray}
When the POT threshold is applied, performance drops to 
\begin{eqnarray}
    \{\overline{F1}^{\text{A}}, \overline{P}, \overline{R}\} &=& \{0.594, 0.635, 0.802\}\nonumber\\
    \{\overline{F1}^{\text{M}}, P^{\text{M}}, R^{\text{M}}\} &=& \{ 0.710, 0.635, 0.802\}\nonumber\\
    \{\overline{F1}^{\text{m}}, P^{\text{m}}, R^{\text{m}}\} &=& \{0.464, 0.320, 0.848\}\nonumber,
\end{eqnarray}
which reflects the higher number of false positives when using POT, in comparison with GS. Also, there is a substantial increase in the variability across machines $\sigma_{\text{machines}} = 0.343$ but there is no  variability across runs as, in contrast to OmniAnomaly, PCA is a deterministic model. It is clear that PCA achieves lower precision than OmniAnomaly (Table~\ref{tab:omnianomaly_smd}), which reduces the resulting F1-score. However, its higher recall indicates that PCA detects more anomalies, albeit at the cost of a larger number of false positives.

\begin{figure}[h]
    \centering
    
    \begin{subfigure}{0.43\textwidth}
        \centering
        \includegraphics[width=\linewidth]{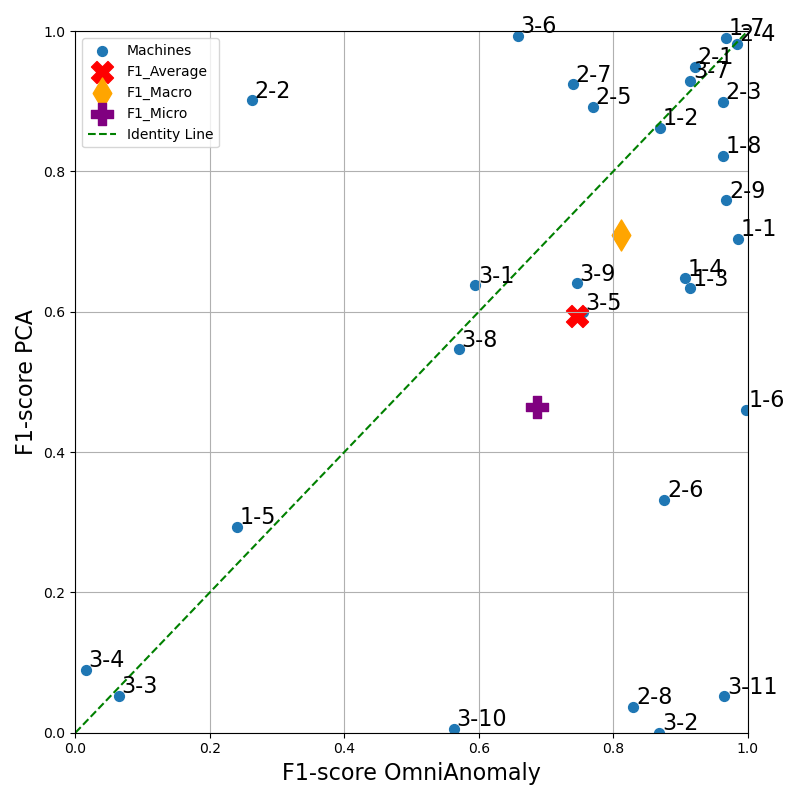}
        \caption{}
        \label{fig:sub1_PA}
    \end{subfigure}
    \hfill
    \begin{subfigure}{0.43\textwidth}
        \centering
        \includegraphics[width=\linewidth]{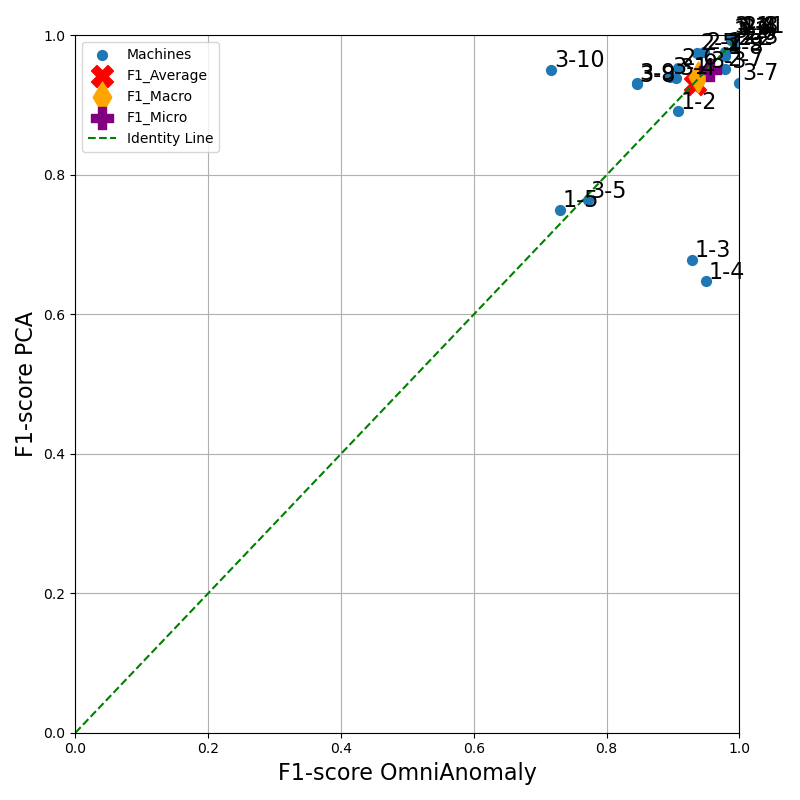}
        \caption{}
        \label{fig:sub2_PA}
    \end{subfigure}
    
    
    \begin{subfigure}{0.43\textwidth}
        \centering
        \includegraphics[width=\linewidth]{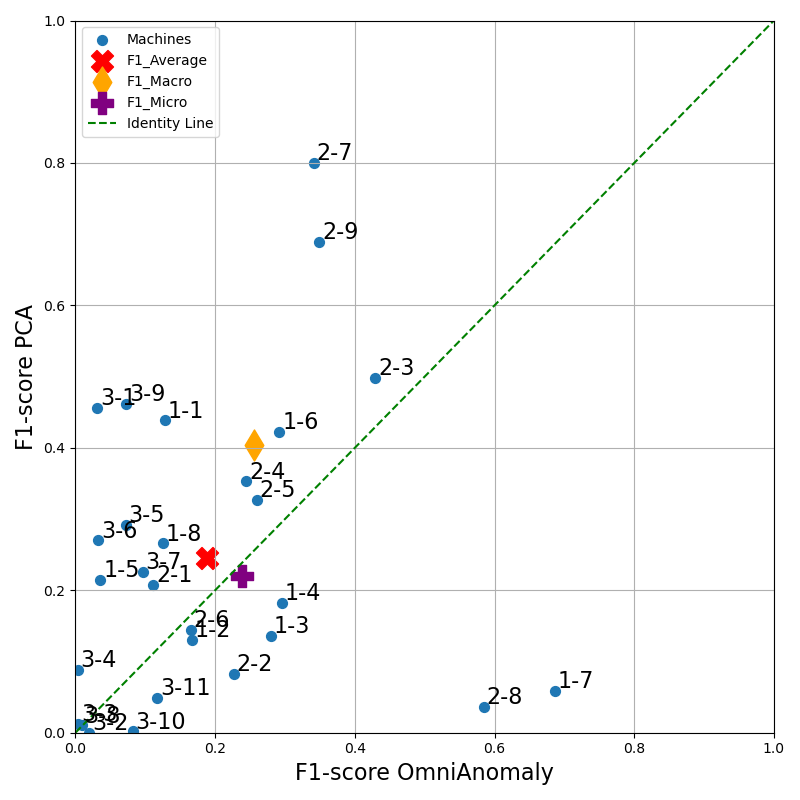}
        \caption{}
        \label{fig:sub3_PA}
    \end{subfigure}
    \hfill
    \begin{subfigure}{0.43\textwidth}
        \centering
        \includegraphics[width=\linewidth]{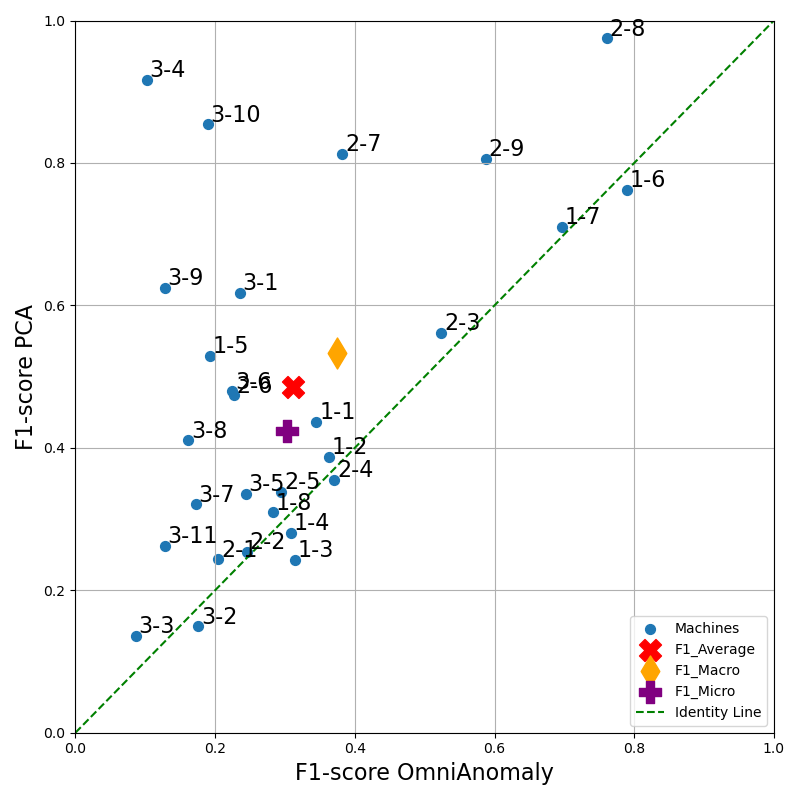}
        \caption{}
        \label{fig:sub4_PA}
    \end{subfigure}
    
    \caption{Comparison between OmniAnomaly and PCA across all machines, (a) POT with PA, (b) GS with PA, (c) POT without PA and (d) GS without PA. Aggregated F1-scores \{$\overline{F1}^{\text{A}}, \overline{F1}^{\text{M}}, \overline{F1}^{\text{m}}$\} represented by \{$\pmb{\times}, \blacklozenge, \pmb{+}$\} symbols.}
    \label{fig: dispersion}
\end{figure}

Figure~\ref{fig: dispersion} provides a machine-level comparison between aggregated F1-scores for OmniAnomaly and PCA under the two thresholding strategies (POT and GS), with and without point-adjustment (PA) evaluation protocols. Figure \ref{fig:sub1_PA} shows that most machines lie near the identity line, indicating comparable performance between both methods. However, some machines (e.g. 2-8, 3-2 and 3-11) achieve F1-scores above 0.80 with OmniAnomaly but near-zero scores with PCA, while others (2-2) show the opposite trend. With the GS threshold  (Fig. \ref{fig:sub2_PA}), results are less dispersed, consistent with the lowest $\sigma_{\text{mach.}}$. Most machines scored similarly for both methods, except 1-3 and 1-4, which still favor OmniAnomaly.
Considering the point-wise evaluation (Figs. \ref{fig:sub3_PA} and \ref{fig:sub4_PA}), the results present a much wider spread and lower scores overall. Using GS, 
\begin{eqnarray}
    \{\overline{F1}^{\text{A}}, \overline{F1}^{\text{M}}, \overline{F1}^{\text{m}}\}&=&\{0.312, 0.374, 0.303\}\nonumber\\
    \{\overline{F1}^{\text{A}}, \overline{F1}^{\text{M}}, \overline{F1}^{\text{m}}\}&=&\{0.485, 0.533, 0.424\}\nonumber
\end{eqnarray}
respectively for OmniAnomaly and PCA while, using POT, the values decrease to \{0.188, 0.255, 0.238\} and \{0.245, 0.404, 0.220\}, respectively. Thus, PCA outperforms OmniAnomaly across most machines regardless of the thresholding strategy. With the POT threshold, only machines 2-8 and 1-7 perform substantially better for OmniAnomaly. With GS, machines such as 3-4, 3-10, and 2-7 reach F1-scores above 0.80 with PCA, while remaining below 0.40 with OmniAnomaly. Overall, these results suggest that performance differences are driven primarily by the choice of thresholding strategy and evaluation protocol, rather than the model itself. The PA protocol inflates scores for both methods and hides meaningful differences, while point-level evaluation exposes true performance and machine-level variability.

In addition to PCA, the arithmetic mean of the standardized metrics was also evaluated as a possible anomaly score (see Eq.~\ref{eq: mean}). Using GS, it achieved 
\begin{eqnarray}
    \{\overline{F1}^{\text{A}}, \overline{P}, \overline{R}\} &=& \{0.886, 0.928, 0.880\}, \sigma_{\text{machines}} = 0.138\nonumber\\
        \{\overline{F1}^{\text{M}}, P^{\text{M}}, R^{\text{M}}\} &=& \{0.904, 0.928, 0.880\},\nonumber\\
    \{\overline{F1}^{\text{m}}, P^{\text{m}}, R^{\text{m}}\} &=& \{0.925, 0.922, 0.928\},\nonumber
\end{eqnarray}
while the performance dropped under POT to 
\begin{eqnarray}
    \{\overline{F1}^{\text{A}}, \overline{P}, \overline{R}\} &=& \{0.567, 0.520, 0.808\}, \sigma_{\text{machines}} = 0.335\nonumber\\
            \{\overline{F1}^{\text{M}}, P^{\text{M}}, R^{\text{M}}\} &=& \{0.633, 0.928, 0.880\},\nonumber\\
    \{\overline{F1}^{\text{m}}, P^{\text{m}}, R^{\text{m}}\} &=& \{0.465, 0.323, 0.828\}.\nonumber
\end{eqnarray}

Without PA, the F1-scores $\{\overline{F1}^{\text{A}},\overline{F1}^{\text{M}},\overline{F1}^{\text{m}}\}$ decrease to $\{0.414,0.448, 0.420\}$ with GS and $\{0.217,0.353, 0.214\}$ with POT. Provided the competitive performance of such a simple arithmetic mean relative to PCA and even OmniAnomaly raises questions about the discriminative power of the SMD dataset. The anomalies may not be sufficiently challenging to distinguish models of very different complexity, consistent with the criticism raised by Wu and Keogh (2021)~\cite{Wu2021}. Furthermore, the high variability across machines (e.g. OmniAnomaly: $\sigma_{\text{machines}} = 0.283$, PCA: $\sigma_{\text{machines}} = 0.343$) suggests that some machines contain trivial anomalies while others are significantly harder to detect. This observation further supports the importance of machine-level analysis, which remains uncommon in current anomaly detection research. Therefore, the results point out that evaluation protocols and dataset characteristics may have a stronger impact on reported performance than model complexity.

\section{Conclusion}

This paper evaluated OmniAnomaly and PCA on the SMD dataset and highlighted several concerns regarding the evaluation and comparison of anomaly detection models. Using the original implementation of OmniAnomaly ensured a fair and reproducible setup, although the exact results reported in the original work could not be reproduced due to the stochastic nature of the model. The analysis also showed that aggregated metrics, such as average F1-scores, may mask substantial performance variability across machines, indicating that machine-level evaluation should become standard practice for heterogeneous datasets like SMD. The results further showed that PCA, despite its simplicity, achieves competitive performance compared to OmniAnomaly and occasionally surpasses it, particularly without point-adjustment. This suggests that part of the reported performance of deep learning models may depend not only on the model itself but also on post-processing choices such as thresholding strategies and evaluation protocols. Overall, these findings highlight the need for more robust evaluation practices and a more critical assessment of the role of deep learning models in time series anomaly detection.





\begin{credits}
\subsubsection{\ackname} 
This work was supported by IEETA (\url{https://www.ieeta.pt/}, UID/00127)
through funding from the Portuguese Foundation for Science and Technology
(FCT, \url{https://www.fct.pt/}, \url{https://ror.org/00snfqn58}),
within the scope of the R\&D grants UID/00127/2025
(\url{https://doi.org/10.54499/UID/00127/2025})
and UID/PRR/00127/2025. B.A. acknowledges the individual PhD scholarship
(ref. 2024.00557.BD), funded by national funds through FCT.

\subsubsection{\discintname}
The authors have no competing interests to declare that are relevant to the content of this article.
\end{credits}
%
%
%
\bibliographystyle{splncs04}
\bibliography{references}
%




\end{document}